# Deep Inception-Residual Laplacian Pyramid Networks for Accurate Single Image Super-Resolution

Yongliang Tang, Weiguo Gong, *Member*, *IEEE*, Xi Chen, and Weihong Li

*Abstract*—With exploiting contextual information over large image regions in an efficient way, the deep convolutional neural network has shown an impressive performance for single image super-resolution (SR). In this paper, we propose a deep convolutional network by cascading the well-designed inception-residual blocks within the deep Laplacian pyramid framework to progressively restore the missing high-frequency details of high-resolution (HR) images. By optimizing our network structure, the trainable depth of the proposed network gains a significant improvement, which in turn improves super-resolving accuracy. With our network depth increasing, however, the saturation and degradation of training accuracy continues to be a critical problem. As regard to this, we propose an effective two-stage training strategy, in which we firstly use images downsampled from the ground-truth HR images as the optimal objective to train the inception-residual blocks in each pyramid level with an extremely high learning rate enabled by gradient clipping, and then the ground-truth HR images are used to fine-tune all the pre-trained inception-residual blocks for obtaining the final SR model. Furthermore, we present a new loss function operating in both image space and local rank space to optimize our network for exploiting the contextual information among different output components. Extensive experiments on benchmark datasets validate that the proposed method outperforms existing state-of-the-art SR methods in terms of the objective evaluation as well as the visual quality.

*Index Terms*—Single image super-resolution, Convolutional neural networks, Laplacian pyramid framework, Local rank space.

## I. INTRODUCTION

Single image super-resolution (SISR) aims to obtain the visually pleasing high-resolution (HR) image from a single low-resolution (LR) image generating by the low-cost imaging system and the limited environment condition. Since the obtained HR images often preserve important details and critical information for later image processing, analysis and interpretation, SISR is topic of great interest in digital image processing and is widely applied to various civilian and military field such as video surveillance [2], medical imaging [3], face recognition [4], satellite imaging [5] and etc.

Manuscript submitted November 10, 2017. This work was supported Key Projects of the National Science and Technology Program (2013GS500303), the Key Science and Technology Projects of CSTC (CSTC2012GG-YYJSB40001, CSTC2013-JCS F40009) and the National Natural Science Foundation of China (61105093)

Yongliang Tang, Weiguo Gong, Xi Chen, and Weihong Li are with Key Lab of Optoelectronic Technology and System of Education Ministry, Chongqing University, Chongqing 400044, China (e-mail: 20150801013@cqu.edu.cn；wggong@cqu.edu.cn；september14@cqu.edu.cn；weihongli@cqu.edu.cn).

SISR problem usually assumes the observed LR image is to be a non-invertible low-pass filtering, down-sampling and noise version of HR image. For this reason, SISR is a highly ill-posed problem. To handle the ill-posed nature in the image SR reconstruction, a variety of methods have been developed in recent years. These methods can be classified into interpolation-based, reconstruction-based, and learning-based methods. The interpolation-based methods, such as bicubic interpolation [6], edge-guided interpolation [7], and nearest neighbor interpolation [8], typically adopt fixed-function kernels or structure-adaptive kernels to estimate the unknown pixels in the HR grid. Although the interpolation-based methods can reconstruct HR images in a very simple and effective way, they are prone to yield overly smooth edges and produce blurring details. Therefore, the reconstructed results are unsatisfactory to the practical applications.

Reconstruction-based SR methods usually introduce certain image priors or constraints between the downsampling of the reconstructed HR image and the original LR images to tackle the ill-posed problem of image SR. These image priors include edge-directed priors [9], gradient profile priors [10], Bayesian priors [11], and nonlocal self-similarity priors [12]. Although this kind of SR methods is particularly effective to preserve geometric structure and to suppress ringing artifacts, it fails to add sufficient novel high frequency details to the reconstructed HR image and is limited in reconstructing the visual complexity of real images, especially for large scaling factor.

Learning-based or example-based methods try to learn mappings from millions of co-occurrence LR-HR example image, and then use the learned mappings to reconstruct the desired HR images. Since the information on example images is exploited in an efficient way, learning-based methods have achieved convincing performance in predicting the high-frequency details lost in HR images. In recent years, the learning-based SR methods have become the research focus in image processing and a variety of algorithms have been proposed including regression-based methods, sparse coding based methods and convolutional neural network (CNN) based methods.

Regression-based methods exploit various regression models to establish the mappings between the LR and HR image spaces. For example, He and Siu [13] utilize Gaussian process regression (GPR) to predict each pixel of the HR image by its neighbors. Timofte et al. [14] propose an anchored neighborhood regression (ANR) approach which uses ridge regression to learn exemplar neighborhoods offline and use these neighborhoods to precompute the mapping between the LR and HR image spaces. More recently, they further propose an improved variant of ANR [15] namely, A+, to improve the accuracy by combining the best qualities of ANR and simple functions. Some other typical methods have also been proposed including support vector regression (SVR) [16], kernel ridge regression (KKR) [17] and structured output regression machine (SORM) [18]. Since the regression-based methods require representing the complicated structures in generic images, the reconstructed HR images usually contain blurring details and ringing artifacts. To alleviate the problem in the regression-based methods, researchers have proposed another kind of learning-based methods that attempt to capture the mappings between the LR and HR images in sparse coding space. Yang et al. [19] presume that the LR and HR image patches share the same sparse code with respect to their own dictionaries and jointly train a compact dictionary pair to imitate the mappings between two image spaces. However, the joint dictionary learning strategy cannot guarantee the co-occurrence of the sparse codes of two image spaces due to the insufficient constraints in the reconstruction phase. To alleviate the inconsistent problem, some sparse coding based methods introduce a variety of image priors or constraints into the reconstruction model, including nonlocal self-similarity prior, [20], regression constraints [21], geometric structure constraints [22], [23] and etc. Although the reconstruction accuracy of the coding based methods have attained impressive improvement over other SR methods, the computational cost is quite high due to the problem of L1 and L2 norms optimization.

Recently, some fast and high-performance models have been successfully applied in image SR. Among them, the CNN-based methods has drawn considerable attention due to its simple structure and excellent reconstruction quality. Dong et al. [24] firstly show that CNN can be used to image SR and train a three-layer convolutional network to learn the mapping from LR image patch space to HR one. Then, Wang et al. [25] introduce the domain knowledge of sparse coding into deep CNN and train a cascade network (SCN) to upsample input images to the desired HR images progressively. Kim et al. [26] propose a highly accurate SR method by using a very deep convolutional network (VDSR) inspired by VGG-net used for ImageNet classification [35]. Lei et al. [33] propose a deep Laplacian pyramid network and further increase the network depth to 27 convolutional layers, which greatly improves the accuracy and speed. In [34], Ledig et al. present a generative adversarial networks framework to generate plausible-looking natural HR images with high perceptual quality. While improving the quality of reconstructed HR images by increasing the networks depth, researchers are also working to achieve real-time single image and video SR reconstruction. Kim et al. [27] propose a shallow network with deeply recursive layers (DRCN) to reduce the number of parameters. ESPCN network [28] use an efficient sub-pixel convolutional layer to replace the bicubic up-sampling operation and extracts features in the LR image space. FSRCNN network [29] adopts a similar idea and uses an hourglass-shaped CNN with more layers but fewer parameters to accelerate SR reconstruction.

Although the above CNN-based SR methods have achieved impressive performance in the reconstruction quality as well as efficiency, they still have deficiency to be addressed. The main limitation for the existing CNN-based methods is always trying to cascade small filters many times in a deep network structure to increase the network depth and improve the reconstruction accuracy, which ignores the influence of network topology structure and leads to the saturation and degradation problem of training accuracy. Besides that, the loss function for most CNN-based methods is the average mean squared error (MSE) which only concerns the overall difference between network output images and corresponding label images. Therefore, the reconstructed SR images have the blurring problem of image small scale structures.

To practically resolve the aforementioned issues, in this paper, we propose a new CNN-based single image SR method, which introduces the well-designed inception-residual blocks into a deep Laplacian pyramid framework to gradually reconstruct the desired HR images from the observed LR input images. Although the trainable depth of the proposed network gains an obvious improvement by optimizing the network structure, the saturation and degradation of training accuracy continues to be a critical problem. In order to reduce our network training difficulty and accelerate the convergence speed, we present an effective two-stage training strategy. In the 1st stage, we firstly downsample the HR images to the size of each pyramid level features step by step, and then the obtained images with different resolution are gradually used to train the inception-residual blocks of the corresponding pyramid level from our network input to output. Inspired by VDSR [26], we also use extremely high learning rates enabled by gradient clipping to accelerate the inception-residual blocks convergence in the each pyramid level. To reduce the training time in the 1st stage, we will cut down the backward computation of the anterior pyramid level and only train the inception-residual blocks in the current pyramid level. When all the inception-residual blocks in each pyramid level are trained, we use the original HR training images to fine-tune the pre-trained inception-residual blocks for obtaining the final SR network model. Since the different components of high-frequency information are lost in different downsampling stages of the HR images and we only learn one stage downsampled images in the inception-residual blocks of each pyramid level, the proposed model can gradually restore the high-frequency information of HR images. Furthermore, we propose a new loss function to alleviate the blurring problem of image small scale structures caused by the average MSE loss function that is used for many CNN-based SR methods. The proposed loss function can optimize our network in both image space and local rank space for exploiting the contextual information among different output components during the network training.

The contributions of this paper can be summarized as follows:

1) Inspired by the Inception and ResNet networks, we propose a very deep convolutional neural network within the Laplacian pyramid framework to gradually reconstruct the desired HR images.

2) To reduce the training difficulty in a very deep networks, we introduce an effective two-stage training strategy to progressively train our network by utilizing the different components of high-frequency information.

3) We propose a new loss function to exploit the contextual information among different output components of the

proposed network.

The remainder of the paper is organized as follows. Section 2 briefly reviews the related work that is important to our image SR method. Section 3 introduces our network structure, the new loss function and the implementation and network training in details. Experimental results and relevant discussions are shown in Section 4 and the conclusions of this paper are given in Section 5.

## II. RELATED WORK

Numerous methods have been proposed to solve single image SR problem. In this section, we focus our discussion on recent CNN-based methods.

### A. Convolutional neural network for SR

In general, the observed LR images can be seen as a degraded product of HR images, which can be generally formulated as [36]

$$y = DHx + v \tag{1}$$

where $x$ and $y$ represent the original HR and observed LR image respectively, $D$ is the downsampling operator, $H$ is the blurring filter, and $v$ represents the additive noise. In view of the above, it is a typical multi-output regression problem to reconstruct a HR image $x$ from the corresponding LR image $y$. Inspired by the promise performance of convolutional neural networks in classification and regression tasks, Dong et al. firstly propose the CNN-based SR method, namely Super-Resolution Convolutional Neural Network (SRCNN). In SRCNN, the end-to-end mapping function $F$ used for reconstructing the desired HR image can be learn by minimizing the following loss function

$$L(\Theta) = \frac{1}{N} \sum_{i=1}^{N} \|F(y_i; \Theta) - x_i\|^2 \tag{2}$$

where $\Theta = \{W_1, W_2, W_3, B_1, B_2, B_3\}$ is the filter and bias of convolutional layers in SRCNN, $x_i$ and $y_i$ represent the HR and LR image patch respectively and $N$ is the number of training samples.

Since SRCNN only uses three convolutional layers (patch extraction/representation layer, non-linear mapping layer and reconstruction layer), it is limited for the contextual information used for reconstructing HR images. To exploit more contextual information in a large image regions, Kim et al. [26] propose a deep CNN-based SR method (VDSR) by cascading $3 \times 3$ convolutional layer many times. Since VDSR utilize the residual-learning to exploit the similarity between input and output image, the loss function used for training the mapping function $F$ is formulated as

$$L(\Theta) = \frac{1}{n} \sum_{i=1}^{n} \|F(y_i; \Theta) - r_i\|^2 \tag{3}$$

where $r_i = y_i - x_i$ is the residual image, $\Theta = \{W_1, \cdots, W_d, B_1, \cdots, B_d\}$, $d = 20$ is the depth of network. Although VDSR significantly improves the reconstruction quality, the restoration of finer texture details is still a challenging problem. To resolve this problem, Ledig et al. [34] propose a perceptual loss function which consists of an adversarial loss and a content loss

$$L(\Theta) = \frac{1}{n} \sum_{i=1}^{n} \|\phi(G_\theta(y_i)) - \phi(x_i)\|^2 - \log D_\theta(G_\theta(y_i))$$
$$+ \|\nabla G_\theta(y_i)\| \tag{4}$$

where $\phi(\cdot)$ is the feature representations, $D_\theta(G_\theta(y_i))$ is the estimated probability that the reconstructed HR image is a natural images and $\nabla G_\theta(y_i)$ is a regularizer based on the total variation to encourage spatially coherent solutions. Among the above methods, the HR images reconstruction is perform in one upsampling step: bicubic interpolation pre-processing or transposed convolution in network output layer. Methods with bicubic interpolation usually leads to high computation complexity since the convolutional operation are applied on the upsampled HR images. On the contrary, although the transposed convolution methods reduce the computational cost, the accuracy of the reconstructed HR images may be affected by the transposed convolutional layer, especially for large scaling factor (e.g.$\times$ 4). Accordingly, some of the CNN-based SR methods introduce Laplacian pyramid framework to strike a balance between reconstruction accuracy and computational cost.

### B. Laplacian pyramid framework

The Laplacian pyramid framework has been used in a wide range of applications, such as edge-aware filtering [30], image blending [31], semantic segmentation [32], and etc. In the image SR, Lei et al. [33] first propose a deep network based on the Laplacian pyramid framework (LapSRN) to reconstruct the HR images. In LapSRN, the network consists of multiple pyramid level and each level is cascaded by a transposed convolutional layer. At each pyramid level, the sub-network is constructed with the same convolutional layers and has its loss function and the corresponding label image $x_s$ downsampled from ground truth HR image with bicubic interpolation. Accordingly, the overall loss function for LapSRN is defined as

$$L(\Theta) = \frac{1}{n} \sum_{i=1}^{n} \sum_{s}^{L} \rho\left((F(y^i; \Theta) + y_s^i) - x_s^i\right) \tag{5}$$

where $L$ is the number of pyramid level, $y_s$ is the upsampled image from the input LR image $y$ in the pyramid level $s$ and $\rho(\cdot)$ is the Charbonnier penalty function used to handle outliers. Since the transposed convolutional layer can merely upsample the images or features with a fixed integer scaling factor, it is limited to the network structure and scaling factors of LapSRN. To super-resolve the input LR image with a more flexible scaling factors, Zhao et al. [37] propose a gradual upsampling network (GUN) for image SR, which use the bicubic interpolation to upsample the features in the forward computation and downsample the gradients in the backward computation during training. However, the proposed method differs from the existing CNN-based SR methods in three aspects.

First, although the proposed method also adopts Laplacian pyramid framework, the structure of our network is obviously different from the existing CNN-based SR methods. LapSRN uses the same structure of convolutional layer to form the sub-network in each pyramid level. In contrast, our network use well-designed inception-residual blocks in each pyramid level to extract features from different scale receptive field. By optimizing the network structure, the trainable depth of the proposed network gains an impressive improvement. Second, we adopt a new two-stage training strategy to further solve the saturation and degradation problem in a very deep networks. With the pre-training for the sub-network in each pyramid level and the fine-tuning for whole network, the reconstruction accuracy for our network has been further improved. Finally, the loss function for most existing CNN-based SR methods primary concerns the overall difference between predicted images and corresponding label images. On the contrary, the proposed loss function also pay attention to the relationship between the each pixels and its neighbors in the network output image.

## III. PROPOSED METHOD

In this section, we describe the design methodology of the proposed convolutional network for single image super-resolution, the optimization using the proposed loss functions with deep supervision, and the details for network training.

### A. Network architecture

As shown in Figure 1(a), the proposed method uses the Laplacian pyramid framework to construct the proposed network model. Our network takes an observed LR image $y$ as input and gradually predicts the high-frequency details of the desired HR image $x$ at each pyramid level. Since each pyramid level of the proposed network has similar structures, we only describe one pyramid level network in details.

In Figure 1 (b), we show the network structure of one pyramid level. For brevity, the activation (ReLU) and batch normalization (BN) layers are not shown in the sub-network structure. As illustrated in Figure 1 (b), the sub-network consists of the upsampling layer, well-designed residual blocks and output layer. At pyramid level $s$, we first use the upsampling layer to upsample the input features $y_{s-1}$ to the size of the level $s$ features, which can be described as,

$$y_s^0 = D_u(y_{s-1}; \theta) \qquad (6)$$

where $y_{s-1}$ is the features of the pyramid level $s-1$, $y_s^0$ is the upsampled features in the pyramid level $s$ and $D_u(\cdot)$ is the upsampling function. In the proposed network, we use an improved transposed convolution to perform the upsampling processing. Comparing to the traditional transposed convolution which can only upsampling the input with a fixed scaling factor, the improved transposed convolution can upsample the input features to a specified resolution more freely. To obtain an upsampled features with an arbitrary resolution, the improved transposed convolution uses a non-uniform method to insert rows and columns of zero elements before the convolution processing. Figure 2 show the inserted features with zero elements by the improved and traditional transposed convolution. Although the bicubic interpolation also can be used to resize the input to an arbitrary resolution [37], it always cause the loss of high frequency features and gradient information in the forward and backward computation. Thus, the upsampling layer for the proposed network can not only upsample the input features to an arbitrary resolution, but also preserve the high frequency information.

Then, the upsampled features will pass the well-designed residual blocks for recovering the specified high-frequency details in the pyramid level $s$. As shown in Figure (c), the residual block for the proposed network is the combination of the two most recent ideas: Residual connections introduced by He et al. in [38] and the latest revised version of the Inception architecture [39]. In [38], the authors have argued that residual connections are of inherent importance for training very deep architectures. To reap the benefits of the residual approach, Szegedy et al. in [1] use residual connections to replace the filter concatenation stage of the Inception architecture and obtain an outperformance on the ImageNet classification challenge. Inspired by this, we introduce the Inception-residual block architecture to the proposed network. The main technical difference between our residual block and the residual block in the Inception-ResNet is that in the case of the proposed network, we use a new residual connections introduced by He et al. in [40] to enhance the capacity of residual block. Accordingly, the residual block for our network can be formulated as,

$$y_s^r = F(y_s^0; \Theta) + y_s^{r-1} \qquad (7)$$

where $y_s^{r-1}$ and $y_s^r$ are the input and output features of the $r$-th residual block respectively.

After the upsampled features passed all the residual blocks, the output of the residual block is connected to two different layers during the training: (1) a convolutional layer for reconstructing a residual image at the pyramid level $s$, and (2) an upsampling layer for super-resolving the features to the size of the finer pyramid level $s+1$. The reconstructed residual image is then combined (using element-wise summation) with the bicubic interpolated image from input LR image to produce a HR output image $\hat{x}_s$. Finally, the produced image $\hat{x}_s$ with the bicubic downsampled image $x_s$ from the ground truth HR image $x$ will be used for training the sub-network of the pyramid level $s$. After completing the training of the network, we will remove the convolutional layer for reconstructing a residual image at each pyramid level and directly predict the residual image between the bicubic upsampled image from input LR image $y$ and the original HR image $x$. Actually, the entire network of the proposed SR method is a cascade of sub-network with a similar architecture at each pyramid level.

### B. Loss function

For most of the CNN-based SR method, the loss function is usually the mean square error (MSE). Thus, the optimization objective for these SR method is defined as,

$$\min_{\Theta} \frac{1}{N} \sum_{i=1}^{N} \left\| F(y^i; \Theta) - x^i \right\|_2^2 \qquad (8)$$

where $y^i$ and $x^i$ are $i$-th LR and HR image patch pair in the training data, and $F(y^i; \Theta)$ is the predicted HR image patch from $y^i$ using the network with parameters $\Theta$. Since MSE struggles to handle the uncertainty inherent relationship in recovering lost high-frequency details such as small scale structures and texture details, the optimization of MSE encourages finding pixel-wise averages of plausible solutions which are typically overly-smooth and thus have poor perceptual quality [46]. To make reconstructed HR images with realistic texture details and sharp edges, we propose a new loss function by exploiting the contextual information among different pixels of the network output images. The local rank transform is an effectively constraint can be used in preserving texture details and edges of the reconstructed image [47] [47], because the contextual information in a rank window is exploited in an efficient way [49]. For a given image $I$, the definition of local rank transform can be formulated as,

$$\text{LRT}_\delta(I(i,j)) = N_w - \sum_{i_w, j_w} \text{C}(I(i,j) - I(i_w, j_w)) \quad (9)$$

where

$$\text{C}(\cdot) = \begin{cases} 1, & I(i,j) - I(i_w, j_w) > \delta \\ 0, & I(i,j) - I(i_w, j_w) \le \delta \end{cases} \quad (10)$$

$N_w$ is the total number of pixels in the rank window, $\delta$ is a parameter making the local rank suit the noisy image, $I(i,j)$ is the pixel value of image $I$ at the rank window center $(i,j)$, and $(i_w, j_w)$ is the pixel coordinate in the rank window. Considering that the local rank transform for a given pixel can be used for describing the statistical distribution characteristics between the given pixel and around pixels in a rank window, we introduce the local rank into the proposed loss function for restricting the contextual relationship among the given pixel and around pixels. Accordingly, the optimization objective for loss function with the introduced local rank constraint can be defined as,

$$\min_\Theta \frac{1}{N} \sum_{i=1}^N \|F(\boldsymbol{y}^i; \Theta) - \boldsymbol{x}^i\|_2^2 \\ + \beta \left\| \text{LRT}_\delta\left(F(\boldsymbol{y}^i; \Theta)\right) - \text{LRT}_\delta(\boldsymbol{x}^i) \right\|_2^2 \quad (11)$$

where $\beta$ is the weight for the local rank constraint. Since our network adopt the Laplacian pyramid framework and residual learning, the overall loss function is defined as,

$$\text{L}(\Theta) = \sum_{i=1}^N \sum_{s=1}^L k_s \left( \|\boldsymbol{r}_s^i - \boldsymbol{x}_s^i\|_2^2 \\ + \beta \|\text{LRT}_\delta(\boldsymbol{r}_s^i) - \text{LRT}_\delta(\boldsymbol{x}_s^i)\|_2^2 \right) \quad (12)$$

where $\boldsymbol{r}_s^i = F(\boldsymbol{y}^i; \Theta) + \boldsymbol{y}_s^i$, $\boldsymbol{x}_s^i$ is downsampled image from the original images to the label images for the pyramid level $s$ using bicubic interpolation, $\boldsymbol{y}_s^i$ is upsampled image from input LR image, $L$ is the number of pyramid level in the proposed network, and $k_s$ is the weight for the loss function of each pyramid level.

### C. Implementation and training details

#### 1) Training dataset

For fair comparison with the other state-of-the-art methods, we also use 91 images from Yang et al. [19] and 200 images from the training set of BSD500 [41] as our training HR images. To make full use of these images, the data augmentation is adopted in the proposed network training. We augment these images data in three ways: (1) Scaling: each HR image is downscaled by bicubic interpolation with the scaling factor 0.9, 0.8, 0.7, and 0.6. (2) Rotation: each image is rotated with the degree of 90, 180 and 270. (3) Flipping: each image is flipped with horizontal and vertical. Thus, we will obtain $5 \times 4 \times 3 = 60$ times image to form the final HR image training set $\{X\}$. Once these HR image trainings are obtained, we can prepare the training data for the proposed network. To generate the training data, we first use bicubic interpolation to downsample the original HR training images $\{X\}$ with the desired scaling factor $n$ to form the corresponding LR image $\{Y\}$. Then we crop the LR training image into a set of

LR image patches $\{\boldsymbol{y}^i\}_{i=1}^N$ with a stride $k$. The corresponding HR image patches $\{\boldsymbol{x}^i\}_{i=1}^N$ are also cropped with a stride $n \times k$ from the HR images. Actually, the cropped LR/HR image patch pairs $\{(\boldsymbol{y}^i, \boldsymbol{x}^i)\}_{i=1}^N$ are the training data for the proposed network. Since the proposed network adopts zero padding in all convolutional layers to keep the size of the features as the same input of each level, it is necessary to employ the LR image patches with the same size for the different factor networks. Thus, for $\times 2$, $\times 3$ and $\times 4$ networks, the size of LR/HR image patches are set to be $27^2/54^2$, $27^2/81^2$ and $27^2/108^2$, respectively.

#### 2) Training methodology

For the proposed SR method, we utilize the Caffe package [42] with stochastic gradient descent algorithm to train our network on an NVidia Titan GPU. Since our network consists of the coupled sub-networks with its loss functions and the corresponding HR image at each pyramid level, the training of our network becomes a critical problem. To reduce the difficulty of network training, we explore a new two-stage training strategy. First, we train the proposed network one pyramid level by one pyramid level using the training data $\{(\boldsymbol{y}^i, \boldsymbol{x}_s^i)\}_{i=1}^N$ downsampled by bicubic interpolation from the original HR training data $\{(\boldsymbol{y}^i, \boldsymbol{x}^i)\}_{i=1}^N$ at pyramid level $s$. When we train the sub-network of pyramid level $s$, we will remove all the output convolutional layers used for reconstructing the residual image and freeze the weights of the others convolutional layers before the pyramid level $s$. Once all the sub-networks are trained, we fine-tune our final network using the original HR training data $\{(\boldsymbol{y}^i, \boldsymbol{x}^i)\}_{i=1}^N$ at a low learning rate. With this strategy, the training converges much earlier than training all the sub-networks together the beginning.

In addition, we provide parameters used for training our final network. In the first training stage. We use a learning rate of 0.1 for the convolutional layers and 0.01 for the improved transposed convolutional layers. The learning rate will be decayed every two epochs using an exponential rate of 0.94. Since we adopt an extremely high learning rates (0.1) to accelerate the convergence, the gradient clipping is set to be 1 and then is decreased by a factor of 10 every two epochs. For weights initialization, all the filters of the convolutional and transposed layers are initialized with the method described in He et al.[43]. During the fine-tuning, the learning rate for all layers is set to be 0.00045 and decayed by an exponential rate of 0.94 every two epochs. Our first stage of training uses momentum [44] with a decay of 0.9, while our fine-tuned models are achieved using RMSProp [45] with decay of 0.9 and $\epsilon = 1.0$. The batches of size and weight decay are set to 128 and 0.0001 for all the training stage, respectively.

## IV. EXPERIMENTAL RESULTS AND DISCUSSIONS

In this section, we first analyze the contributions of the different components of the proposed network. Then, we compare the proposed method with state-of-the-art SR algorithms on five representative image datasets and show the super-resolution results on the real-world images. Finally, we discuss the computational complexity of the proposed algorithm in terms of reconstructing phase.

In our experiments, we follow the publicly availabel evaluation framework of Timofte et al. [14]. It enables the comparison of the proposed method with many state-of-the-art SR methods in the same setting. The framework first crops pixels near image boundary and enables the size of cropped images with respect to the target scaling factors. Then, considering that the human visual system is more sensitive to details in intensity than color, the framework transforms the cropped images from RGB color space into YCbCr color space and applies bicubic interpolation on the transformed images for obtaining the input LR images in YCbCr space. Finally, this framework applies the SR reconstruction algorithm on the luminance channel and directly upscales the chrominance (Cb and Cr) channels to the desired resolution using bicubic interpolation. Furthermore, SR performance metrics including the peak signal-to-noise ratio (PSNR) and structural similarity (SSIM) are applied to evaluate the objective quality of reconstructed HR images.

### A. Investigation of the proposed network

In this section, we design a set of controlling experiments to analyze the property of the proposed network and confirm the contributions of the different components of our network for the accuracy of image SR reconstitution.

#### 1) Topology structure of the proposed network

To demonstrate the effect of the network structure of our network, we remove all the inception-residual blocks in the deep Laplacian pyramid framework and directly cascade the same convolutional layers many times at each pyramid level. For a fair comparison, the number of convolutional layers and weight parameters for the degraded network is same as the proposed network. Moreover, the proposed network and degraded network all are optimized using MSE loss function from the weights initialized randomly by the same method. Figure 3 shows the convergence curves in terms of PSNR on the Set14 for the scaling factor of 3. The performance of the degraded network (blue curve) is significantly worse than the network with the inception-residual structure. The proposed network (coffee curve) has better convergence accuracy and speed. In Figure 4, we show that the degraded network reconstructs HR image with more blurring edges and details. In contrast, the reconstructed HR images provided by the proposed network contain more clean edges and visual details. In view of the above, the network with the inception-residual blocks in a deep Laplacian pyramid framework is more capable of reconstructing the desired HR images.

#### 2) Loss function

In the proposed method, we present a new loss function for our network to exploit the contextual information among different output components during training. Here we verify the effectiveness of the proposed loss function. For a fair comparison, we use the same training parameters and initialization method to optimize the proposed network with the MSE loss function. As illustrated in Figure 3, the network optimized with our proposed loss function (red curve) requires more iterations to achieve comparable performance with FSRCNN and the convergence curve also fluctuates significantly. This is mainly because that the proposed loss function apply the more stringent constraints among the predicted images and original HR training images. However, the optimized network with the proposed loss function has shown to promise an excellent reconstruction accuracy. In addition, we show the super-resolution results by the proposed algorithm and the proposed network trained with MSE loss function in Figure 4. As shown in Figure 4 (d), the reconstructed HR images by the network trained with MSE loss function contain blurring details and ringing artifacts. In contrast, the proposed method show promising performance in preserving sharpen edges, reconstructing visual details, and suppressing ringing artifacts because the more contextual information among the different components is exploited in an efficient way.

#### 3) Two-Stage training strategy

Since the depth of the proposed network is remarkably increased by cascading more one well-designed inception-residual blocks within the deep Laplacian pyramid network, the convergence accuracy and speed becomes a critical issue during training. In order to alleviate this problem, we adopt a two-stage training strategy to optimize our network. Here we validate the effectiveness of the proposed training strategy by comparing our final model to that of the proposed network trained from the randomly initialized condition. For a fair comparison, we also use extremely high learning rate enabled by adjustable gradient clipping to accelerate the convergence of the randomly initialized network. As shown in Figure 3, the trained network with our proposed two-stage strategy (yellow curve) has better convergence accuracy, and there is no obvious fluctuation in the convergence curve of fine-tuning network. Furthermore, we report the PSNR and SSIM on the Set5 and Set14 for the scaling factor of 3 in Table 1 and show the reconstructed HR images by the randomly initialized network in Figure 4. From the tables, the average PSNR gains of the proposed training strategy are 0.21dB and 0.18dB, respectively. In Figure 4, we also can observe that our final SR model generates HR results with less ringing artifacts and more sharpen edges. It is obvious that the proposed training strategy can further improve the reconstructed HR images quality.

#### 4) Network depth

To demonstrate the effect of network depth, we train the proposed network with different depth and show the trade-off between super-resolving performance and speed in Table 2 and 3. Since our network depth is decided by the number of pyramid level $L$ and the number of inception-residual blocks $d$ on the each pyramid level, we train the proposed network with different $L$ and $d$ to validate the effect of the two parameters independently. To show the effect of the number of pyramid level $L$, we first fix the $d$ as 1 and then train the our network with different number of pyramid level $L = 1, 2, 3, 4, 5$. Table 2 illustrates the average PSNR values and running time on the Set5 and Set14 with different number of pyramid level. In general, deep networks perform better than shallow ones at the expense of increased computational cost. However, the PSNR results increase slowly when $L$ is larger than 4. Accordingly, we choose $L = 2, 4, 6$ for our $2 \times$, $3 \times$ and $4 \times$ SR networks to strike a balance between super-resolving performance and efficiency, respectively.

After we choose the number of pyramid level $L$ for our SR models, we can train our network with the different number of inception-residual blocks $d = 1, 2, 3, 4, 5$ in the pyramid level to validate the effect of the sub-network depth at each pyramid level of the proposed network. In Table 3, we report the average PSNR values and average running time on the Set5 and Set14 with different $d = 1, 2, 3, 4, 5$. As shown in Table 3, it is obvious that the proposed network with more inception-residual blocks at each pyramid level can improve the quality of reconstructed HR images. However, the PSNR values increase slowly at the expense of increased computational cost when the number of inception-residual blocks at each pyramid level is larger than 3. Hence, we choose $d = 3$ for our $2 \times$, $3 \times$ and $4 \times$ SR models for striking a balance between reconstruction quality and execution time, respectively.

### B. Comparisons with the state-of-the-arts

To validate the performance of the proposed method, the SR experiments for different scaling factors ($\times 2$, $\times 3$ and $\times 4$) are performed on all the images in the five representative image datasets Set5, Set14, BSD100, Urban100 and Manga109 [30]. Among these datasets, Set5, Set14 and BSDS100 consist of natural scenes images; Urban100 contains challenging urban scenes images with details in different frequency bands; and MANGA109 is a dataset of Japanese manga. Then, we compare the proposed method with 6 state-of-the-art SR algorithms: FSRCNN [29], SelfExSR [50], SCN [25], VDSR [26], DRCN [27], and LapSRN [33]. All the compared results are reproduced by the codes downloaded from the authors' websites under the same setting with our experiments.

Table 4, 5, 6 shows the average PSNR and SSIM results of reconstructed HR images on the five representative image datasets by the proposed and baseline methods with different scaling factors, respectively ($\times 2$, $\times 3$ and $\times 4$). From these tables, we can see that the proposed method achieves the consistent performance on all the test image datasets. Since the proposed and compared methods are based on the learning methods and the training data are limit, the performance for all methods tends to decline with the increase of test examples. However, the proposed method still performs better than all the compared methods in both PSNR and SSIM. These experimental results suggest that the proposed method can effectively improve the quality of reconstructed HR images.

To assess the visual quality of the proposed method, we show reconstructed HR images by the proposed and compared methods on the images, drawn from B100, Urban100 and Manga109, with different scaling factors ($\times 2$, $\times 3$ and $\times 4$) in Figure 5, 6 and 7, respectively. As shown in these figures, our proposed method accurately reconstructs parallel straight lines and grid patterns such as windows and the stripes on tigers. We observe that methods using the bicubic upsampling for pre-processing generate results with noticeable artifacts. In contrast, our approach effectively suppresses such artifacts through progressive reconstruction and the proposed new loss function.

### C. Super-resolving on real-world images

In this section, we perform our proposed method and some compared methods on the historical photographs with JPEG compression artifacts to demonstrate the super-resolving performance on real-world images. In our experiment, neither the ground-truth HR images nor the downsampling kernels are available. The super-resolved historical images for different upsampling scale factors ($\times 2$, $\times 3$ and $\times 4$) are shown in Figure 8, 9 and 10, respectively. As shown in Figure 9, the proposed method can provides clearer details and sharper edges in the reconstructed HR historical images than other existing state-of-the-art methods.

### D. Computational complexity

For the reality SR applications, the time complexity of the algorithm also needs to be considered owing to the limitation of computing resource. In this section, we discuss the time complexity of the proposed algorithm. For a fair comparison, the proposed and compared methods are conducted on the same platform with Ubuntu 14.04 operating system, 3.5 GHz Intel i7-5960x CPU, 64 GB memory and NVIDIA Titan X GPU. Since the testing codes for FSRCNN are based on CPU implementations, we reconstruct the network of FSRCNN in Caffe with the same weights to measure the execution efficiency on GPU. Fig. 11 show trade-offs between the execution time and PSNR values by the proposed method and the compared methods on Set14 dataset for the scaling factor of 4. Since the proposed method has more complicated network structure and applies more convolutional layers on larger upsampling scale factors, the time complexity of our network increases slightly with respect to the target scaling factors. However, the reconstructed quality of our method still performs favorably against LapSRN, DRCN and other existing SR methods.

## V. CONCLUSION

In this paper, we propose an accurate single image super-resolution method using very deep convolutional neural network within the Laplacian pyramid framework. In the proposed method, our network consists of an input upsampling layer, the well-designed inception-residual blocks on each pyramid level, and an output layer. Since the different components of the missed high-frequency information are gradually recovered by the well-designed inception-residual blocks of each pyramid level, the proposed network can efficiently learn to reconstruct the HR images. In order to alleviate the difficulty of training in a very deep networks, we adopt a simple yet effective two-stage training strategy, in which the different components of high-frequency information are firstly utilized to train gradually the inception-residual blocks of each pyramid level as the initial weights of the proposed network, and then the original HR training images with the more complex high-frequency information are used for fine-tuning our final network model. Furthermore, we present a new loss function for the proposed network, in which our network is optimized in both image space and local rank space to exploit the contextual information among different output components. Experimental results on a number of images, drawn from five representative image datasets, demonstrate that the proposed method can obtain accurate results and performs competitively in comparison to other existing state-of-the-art SR methods.

## REFERENCES

[1] C. Szegedy, S. Ioffe, V. Vanhoucke, A. Alemi. Inception-v4, Inception-ResNet and the Impact of Residual Connections on Learning, arXiv preprint arXiv: 1602.07261, 2016.

[2] L. Zhang, H. Zhang, H. Shen, and P. Li. A super-resolution reconstruction algorithm for surveillance images, Signal Processing, 90 (2010) 848–859.

[3] S. Peled and Y. Yeshurun. Super-resolution in MRI: application to human white matter fiber tract visualization by diffusion tensor imaging. official journal of the Society of Magnetic Resonance in Medicine, 45 (2001) 29–35.

[4] B. K. Gunturk, A. U. Batur, Y. Altunbasak, M. H. Hayes, and R. M. Mersereau. Eigenface-domain super-resolution for face recognition. IEEE Transactions on Image Processing, 12(5):597–606, 2003.

[5] M. W. Thornton, P. M. Atkinson, and D. a. Holland. Sub-pixel mapping of rural land cover objects from fine spatial resolution satellite sensor imagery using super-resolution pixel-swapping. International Journal of Remote Sensing, 27(3):473–491, 2006.

[6] R. Keys, Cubic convolution interpolation for digital image processing, IEEE Trans. Acoust., Speech, Signal Process. 29 (06) (1981) 1153-1160.

[7] L. Zhang, X. Wu, An edge-guided image interpolation algorithm via directional filtering and data fusion, IEEE Trans. Image Process. 15 (08) (2006) 2226-2238.

[8] R. Fattal, Image up-sampling via imposed edge statistics, ACM Transactions on Graphics 26 (03) (2007) 095-103.

[9] Y.-W. Tai, S. Liu, M. S. Brown, S. Lin, Super resolution using edge prior and single image detail synthesis, in Proc. IEEE Conf. Comput. Vis. Pattern Recognit., 2010, pp. 2400–2407.

[10] L. Wang, S. Xiang, G. Meng, H. Wu, C. Pan, Edge-directed single-image super-resolution via adaptive gradient magnitude self-interpolation, IEEE Transactions on Circuits & Systems for Video Technology 23 (08) (2013) 1289-1299.

[11] N. Akhtar, F. Shafait, A. Mian. Bayesian sparse representation for hyperspectral image super resolution, in Proc. IEEE Conf. Comput. Vis. Pattern Recognit., 2015, pp. 3631-3640.

[12] K. Zhang, X. Gao, D. Tao, and X. Li, "Image super-resolution via nonlocal steering kernel regression regularization," in Proc. 20th IEEE Conf. Image Process, 2013, pp. 943–946.

[13] H. He, W.-C. Siu, Single image super-resolution using Gaussian process regression, in Proc. IEEE Conf. Comput. Vis. Pattern Recognit., Jun. 2011, pp. 449–456.

[14] R. Timofte, V. De Smet, L. Van Gool, Anchored neighborhood regression for fast example-based super-resolution, in Proc. IEEE Int. Conf. Comput. Vis., Dec. 2013, pp. 1920–1927.

[15] R. Timofte, V. De Smet, L. Van Gool, A+: Adjusted anchored neighborhood regression for fast super-resolution, in Proc. 12th Asian Conf. Comput. Vis., Nov. 2014, pp. 111–126.

[16] K. S. Ni, and T. Q. Nguyen, Image super resolution using support vector regression, IEEE Transactions on Image Processing, 16(6) (2007) 1596-1610.

[17] K. I. Kim, and Y. Kwon, Single-image super-resolution using sparse regression and natural image prior, IEEE Trans. Pattern Anal. Mach. Intell., 32(06) (2010) 1127-1132.

[18] C. Deng, J. Xu, K. Zhang, D.Tao, X. Gao, and X. Li, Similarity Constraints-Based Structured Output Regression Machine: An Approach to Image Super-Resolution, IEEE Trans Neural Netw Learn Syst., 27(12) (2015) 2472-2485.

[19] J. Yang, J. Wright, T. S. Huang, Y. Ma, Image super-resolution via sparse representation, IEEE Trans. Image Process. 19 (11) (2010) 2861-2873.

[20] W. Dong, L. Zhang, G. Shi, X. Li, Nonlocally centralized sparse representation for image restoration, IEEE Trans. Image Process. 22 (4) (2013) 1620–1630.

[21] W. Yang, Y. Tian, F. Zhou, Consistent coding scheme for single image super-resolution via independent dictionaries, IEEE Transactions on Multimedia, 18 (03) (2016) 313-325.

[22] W. Gong, L. Hu, J. Li, W. Li, Combining sparse representation and local rank constraint for single image super resolution, Information Sciences 325 (2015) 1-19.

[23] W. Gong, Y. Tang, X. Chen, et al. Combining Edge Difference with Nonlocal Self-similarity Constraints for Single Image Super-Resolution, Neurocomputing 249 (2017) 157-170.

[24] C. Dong, C. Loy, K. He, and X. Tang, Image super-resolution using deep convolutional networks, IEEE Trans. Pattern Anal. Mach. Intell., 38 (2) (2016) 295–307.

[25] Z. Wang, D. Liu, J. Yang, W. Han, and T. Huang. Deep networks for image super-resolution with sparse prior. In ICCV, 2015.

[26] Kim, Jiwon, J. K. Lee, and K. M. Lee. Accurate Image Super-Resolution Using Very Deep Convolutional Networks, in Proc. IEEE Conf. Comput. Vis. Pattern Recognit., 2016, pp. 1646-1654.

[27] Kim, Jiwon, J. K. Lee, and K. M. Lee. Deeply-Recursive Convolutional Network for Image Super-Resolution, in Proc. IEEE Conf. Comput. Vis. Pattern Recognit., 2016, pp. 1637-1645.

[28] W. Shi, J. Caballer, F. Huszar, et al., Real-Time Single Image and Video Super-Resolution Using an Efficient Sub-Pixel Convolutional Neural Network, in Proc. IEEE Conf. Comput. Vis. Pattern Recognit., 2016, pp. 1874-1883.

[29] C. Dong, C. C. Loy, and X. Tang. Accelerating the super resolution convolutional neural network. In ECCV, 2016.

[30] S. Paris, S. W. Hasinoff, and J. Kautz. Local laplacian filters: Edge-aware image processing with a Laplacian pyramid. ACM TOG (Proc. of SIGGRAPH), 30(4):68, 2011.

[31] P. J. Burt and E. H. Adelson. The Laplacian pyramid as a compact image code. IEEE Transactions on Communications, 31(4):532 – 540, 1983.

[32] G. Ghiasi and C. C. Fowlkes. Laplacian pyramid reconstruction and refinement for semantic segmentation. In ECCV, 2016.

[33] W. Lai, J. Huang, N. Ahuja, M. Yang, Deep Laplacian Pyramid Networks for Fast and Accurate Super-Resolution. in Proc. IEEE Conf. Comput. Vis. Pattern Recognit., 2017.

[34] Ledig, Christian, et al. "Photo-Realistic Single Image Super-Resolution Using a Generative Adversarial Network." (2016).

[35] K. Simonyan and A. Zisserman. Very deep convolutional networks for large-scale image recognition. In ICLR, 2015.

[36] Yang S, Wang M, Y. Chen, Y. Sun, Single-image super-resolution reconstruction via learned geometric dictionaries and clustered sparse coding, IEEE Trans. Image Process. 21 (09) (2012) 4016–4028.

[37] Y. Zhao, R.Wang, W. Dong, W. Jia, J. Yang, X. Liu, W. Gao. GUN: Gradual Upsampling Network for single image super-resolution, arXiv preprint arXiv: 1703. 04244, 2017.

[38] K. He, X. Zhang, S. Ren, and J. Sun. Deep residual learning for image recognition. arXiv preprint arXiv: 1512.03385, 2015.

[39] C. Szegedy, V. Vanhoucke, S. Ioffe, J. Shlens, and Z. Wojna. Rethinking the inception architecture for computer vision. arXiv preprint arXiv: 1512.00567, 2015.

[40] K. He, X. Zhang, S. Ren, J. Sun. Identity mappings in deep residual networks. in Proc. European Conference on Computer Vision, 2016, pp.630-645.

[41] P. Arbelaez, M. Maire, C. Fowlkes, and J. Malik. Contour detection and hierarchical image segmentation. TPAMI, 33(5):898–916, 2011.

[42] Jia, Y., Shelhamer, E., Donahue, J., Karayev, S., Long, J., Girshick, R., Guadarrama, S., Darrell, T.: Caffe: Convolutional architecture for fast feature embedding. In: ACM MM. (2014) 675–678

[43] K. He, X. Zhang, S. Ren, and J. Sun. Delving deep intorectifiers: Surpassing human-level performance on ImageNet classification. CoRR, abs/1502.01852, 2015.

[44] I. Sutskever, J. Martens, G. Dahl, and G. Hinton. On the importance of initialization and momentum in deep learning. In Proceedings of the 30th International Conference on Machine Learning (ICML-13), volume 28, pages 1139–1147. JMLR Workshop and Conference Proceedings, May 2013.

[45] T. Tieleman and G. Hinton. Divide the gradient by a running average of its recent magnitude. COURSERA: Neural Networks for Machine Learning, 4, 2012. Accessed: 2015-11-05.

[46] M. Mathieu, C. Couprie, and Y. LeCun. Deep multi-scale video prediction beyond mean square error. arXiv preprint arXiv: 1511. 05440, 2015.

[47] W. Gong, L. Hu, J. Li, W. Li, Combining sparse representation and local rank constraint for single image super resolution, Information Sciences 325 (2015) 1-19.

[48] W. Gong, Q. Yi, Y. Tang, W. Li, Multi-layer strategy and reconstruction model with low rank and local rank regularizations for single image super-resolution, Signal Processing Image Communication, 57 (2014) 197-210.

[49] J. Banks, M. Bennamoun, Reliability analysis of the rank transform for stereo matching, IEEE Trans. Syst. Man, Cybernet. Part B 31 (6) (2001) 870–880.

[50] J.-B. Huang, A. Singh, and N. Ahuja. Single image super-resolution from transformed self-exemplars. In CVPR, 2015.

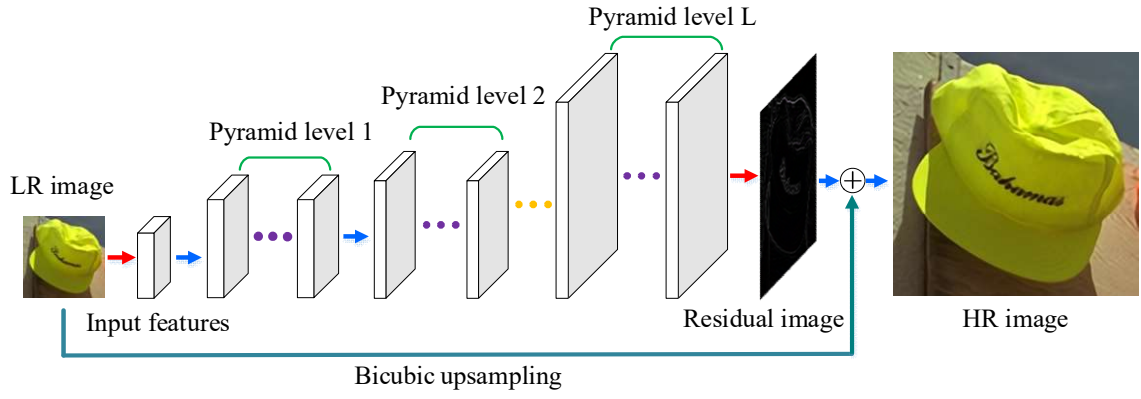

(a)

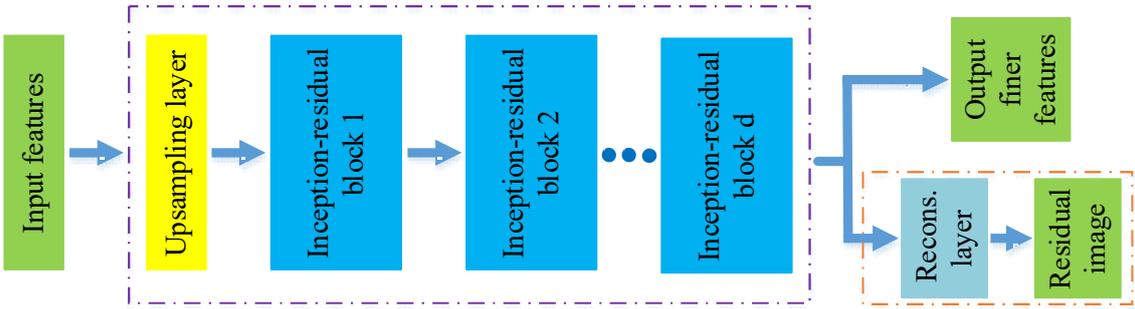

(b)

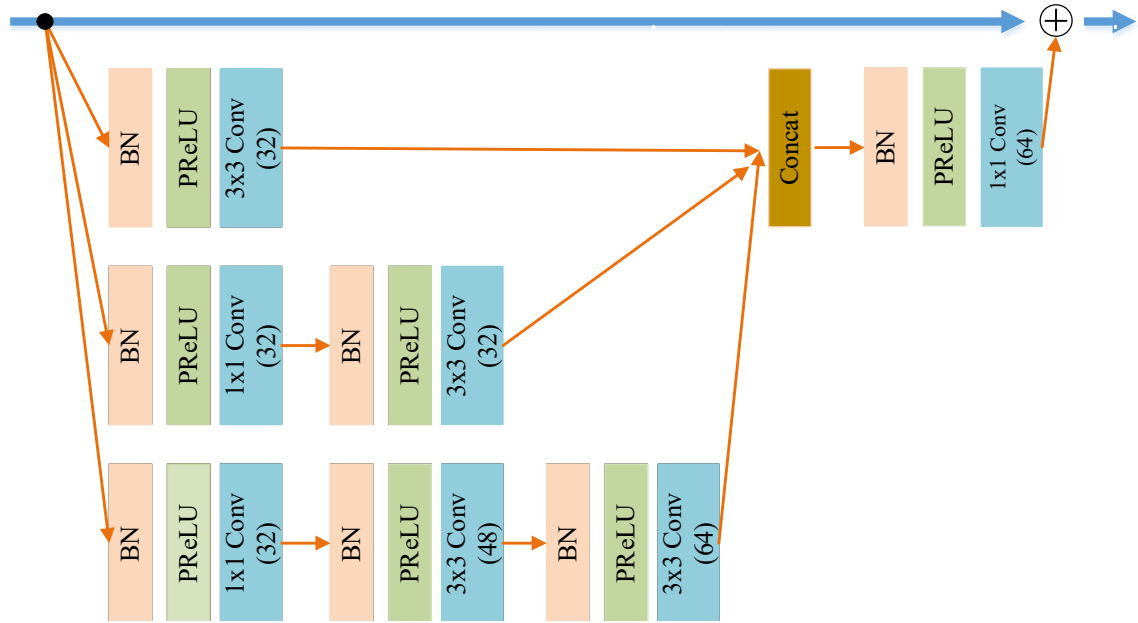

(c)

**Fig.1** Framework of the proposed method. (a) The overall structures of the proposed network. Red arrows indicate convolutional layers, and green arrows denote upsampling operators. (b) The sub-network structures of each Pyramid level. The convolutional layer for reconstructing the residual images (orange double dot line rectangular box) is only activity in the pre-training of each Pyramid level sub-network. (c) The schema for Inception-residual module on the each Pyramid level of the proposed network.

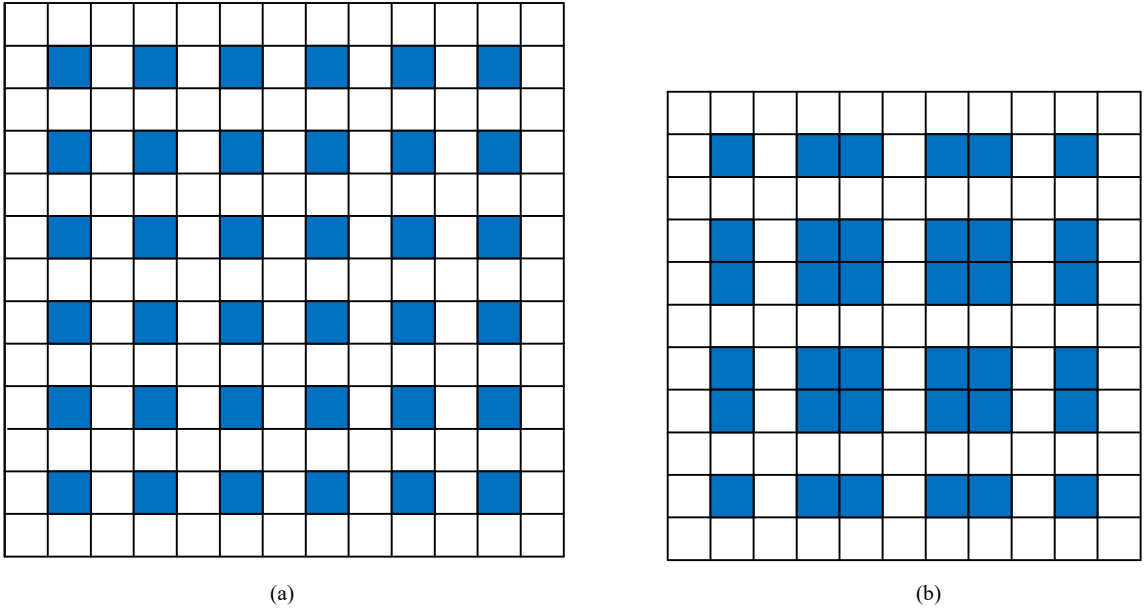

<p style="text-align:center">(a)                                 (b)</p>

**Fig.2** The inserted feature maps with zero elements by the improved and traditional transposed convolutional layers (Blue squares represent the elements of the original feature maps. White squares indicate the inserted zero elements). (a) The inserted feature maps with traditional transposed layers (3x3 transposed, padding: 1, stride: 2, the size of output features: $2 \times (6-1) + 1 \times (3-1) + 1 - 2 \times 1 = 11$). (b) The inserted feature maps with our transposed layers (3x3 transposed, padding: 1, stride: 1.5, the size of output features: $1.5 \times (6-1) + 1 \times (3-1) + 1 - 2 \times 1 = 9$).

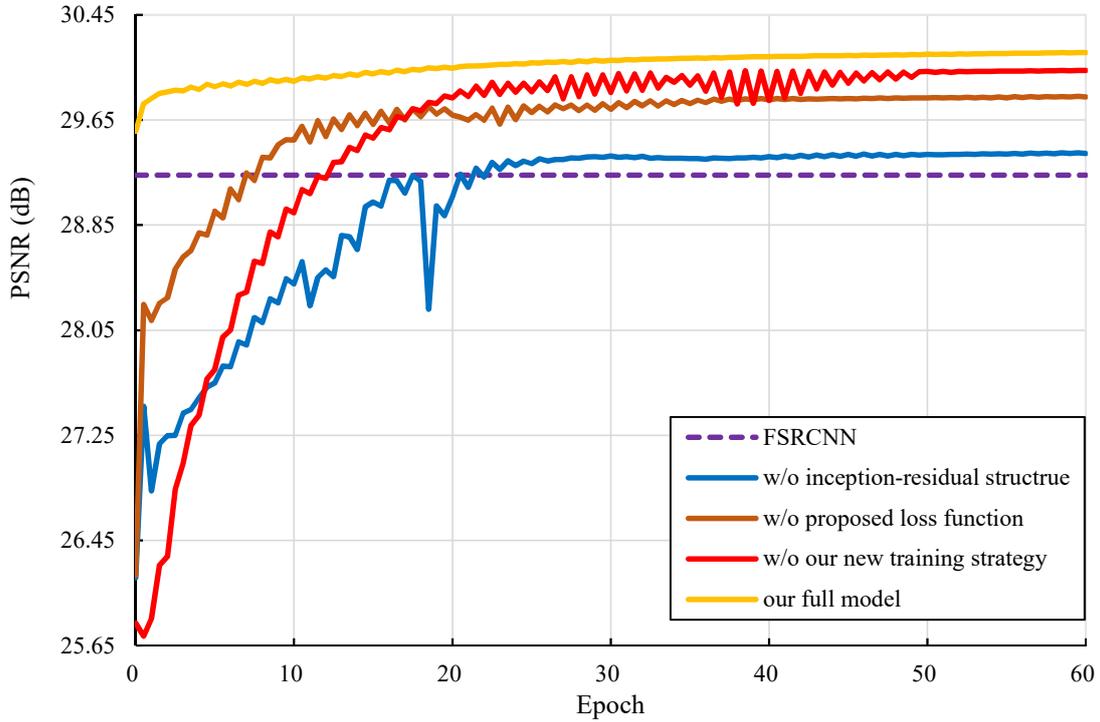

**Fig.3** Convergence analysis on the network structure, loss function and training strategy. The results are obtained on all the images in Set14 with the scale factor 3. Blue curve is the convergence analysis on the network without the inception-residual structure and the proposed loss function. Coffee curve is the convergence analysis for our network with the MSE loss function. Red curve is the proposed network trained with the random initializing method. Yellow curve is the convergence analysis on the fine-tuning of the proposed network.

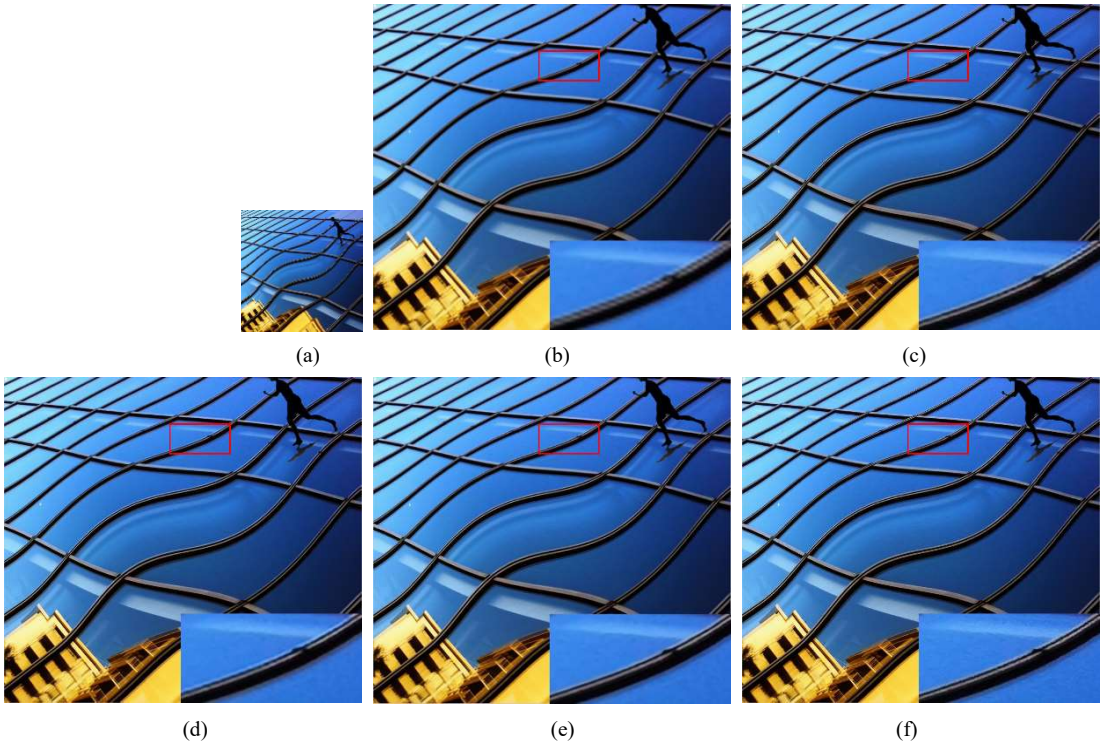

**Fig.4** Visual comparison of reconstructed results by the different components in the proposed network. (scaling factor × 3). (a) input LR image. (b) Bicubic HR image. (c) w/o inception-residual structure. (d) w/o the proposed loss function. (e) the proposed network (full model). (f) ground truth HR image.

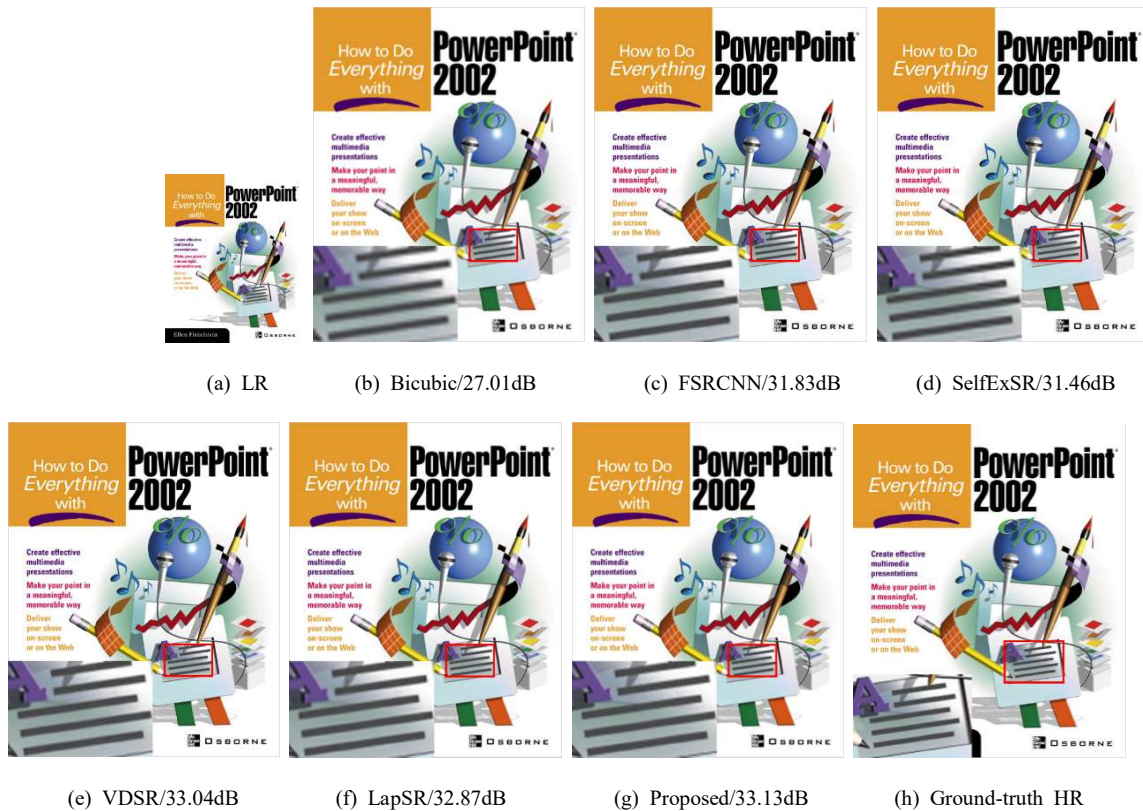

(a) LR      (b) Bicubic/27.01dB      (c) FSRCNN/31.83dB      (d) SelfExSR/31.46dB

(e) VDSR/33.04dB      (f) LapSR/32.87dB      (g) Proposed/33.13dB      (h) Ground-truth HR

**Fig.5** Visual comparison of reconstructed results on "ppt3" image (SET14) by the baseline and proposed methods (scaling factor × 2).

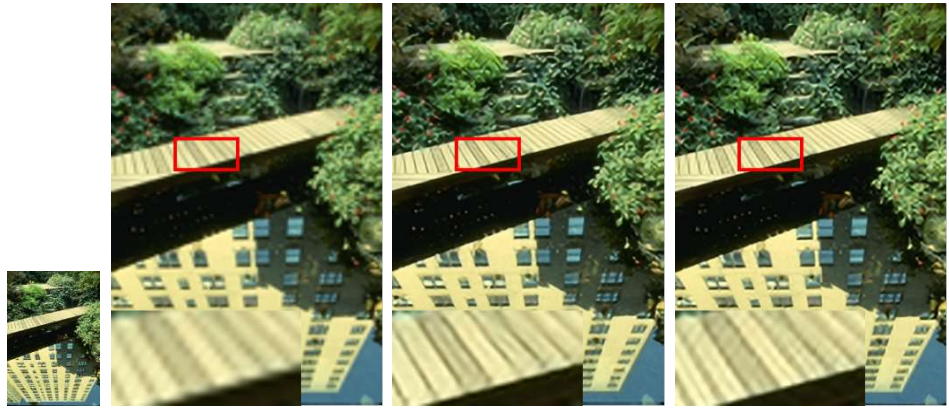

(a) LR     (b) Bicubic/21.96dB     (c) FSRCNN/23.12dB     (d) SelfExSR/22.97dB

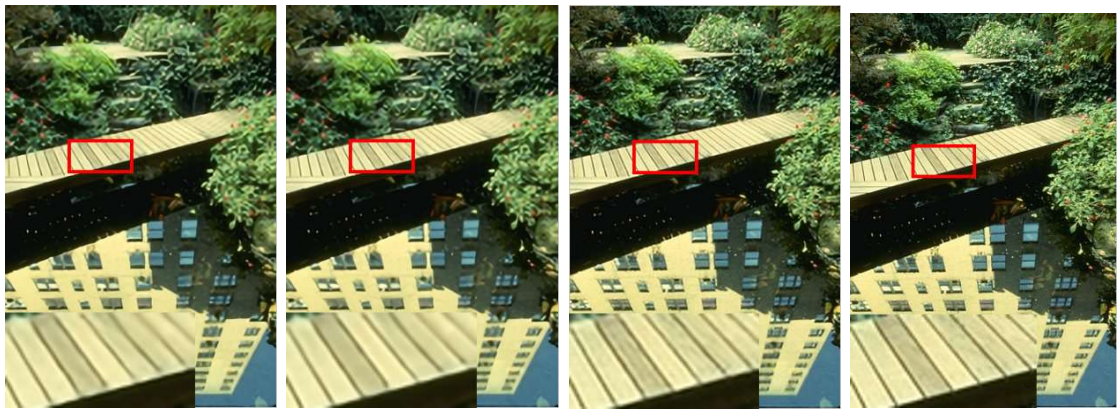

(e) VDSR/23.45dB     (f) DRCN/23.44dB     (g) Proposed/23.62dB     (h) Ground-truth HR

**Fig.6** Visual comparison of reconstructed results on "148026" image (BSD100) by the baseline and proposed methods (scaling factor × 3).

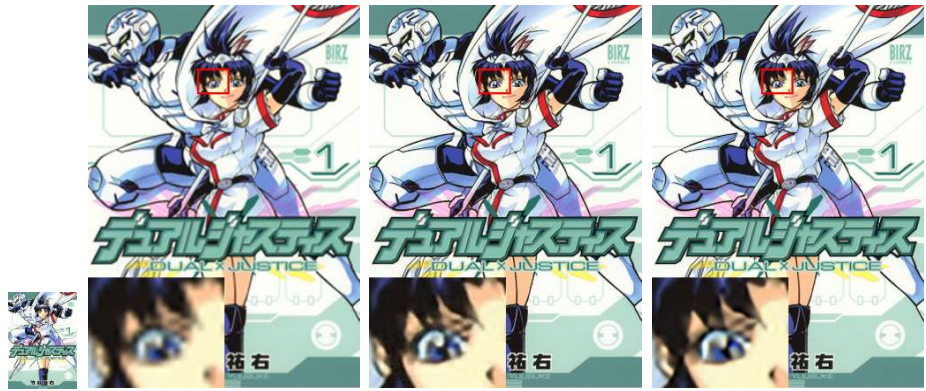

(a) LR     (b) Bicubic/23.58dB     (c) FSRCNN/27.21dB     (d) SelfExSR/26.79dB

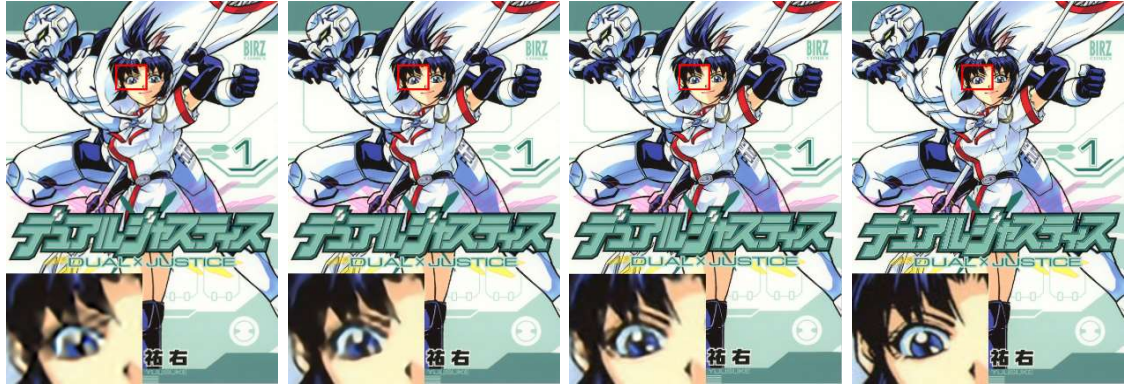

(e) VDSR/28.02dB    (f) LapSR/28.29dB    (g) Proposed/28.45dB    (h) Ground-truth HR

**Fig.7** Visual comparison of reconstructed results on "DualJustice" image (Manga109) dataset by the baseline and proposed methods (scaling factor × 4).

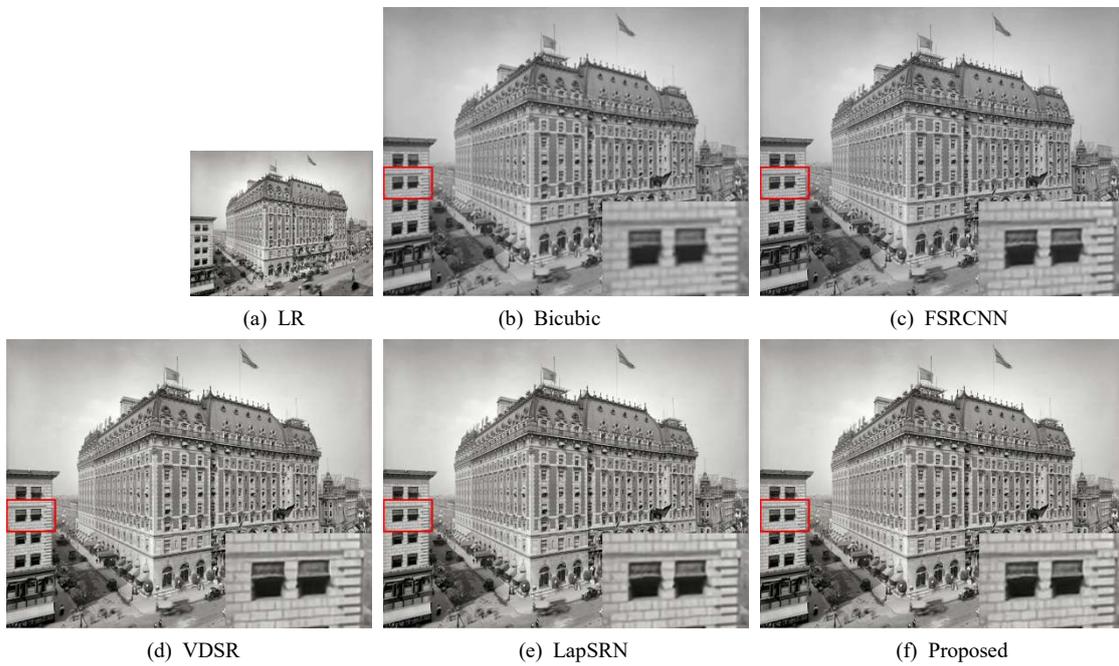

(a) LR    (b) Bicubic    (c) FSRCNN

(d) VDSR    (e) LapSRN    (f) Proposed

**Fig.8** Visual comparison of reconstructed results on real-world historical image by the baseline and proposed methods (scaling factor × 2).

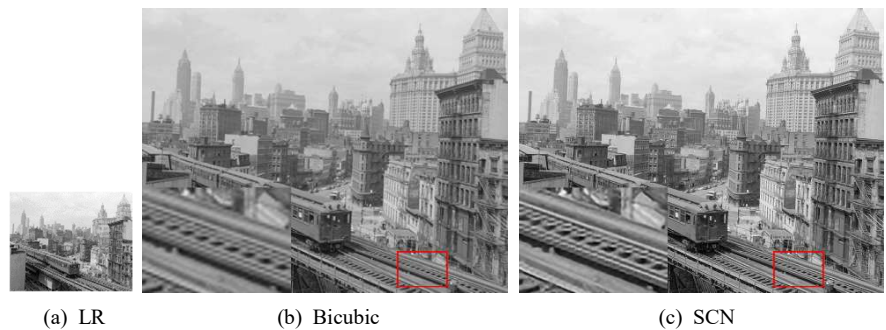

(a) LR    (b) Bicubic    (c) SCN

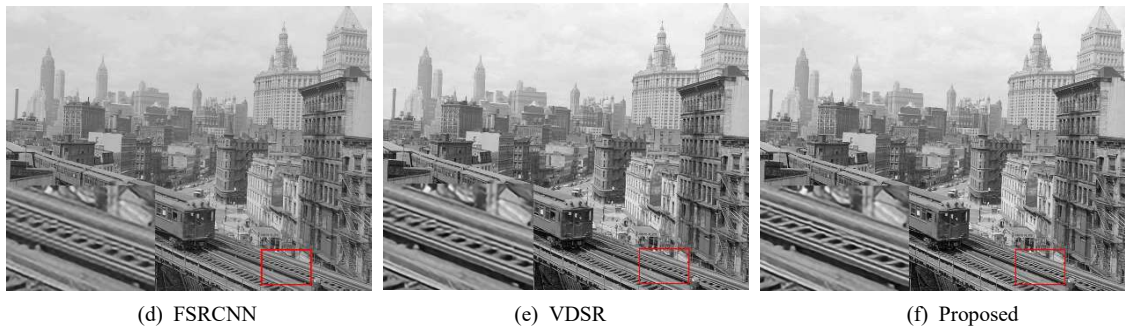

(d) FSRCNN          (e) VDSR          (f) Proposed

**Fig.9** Visual comparison of reconstructed results on real-world historical image by the baseline and proposed methods (scaling factor × 3).

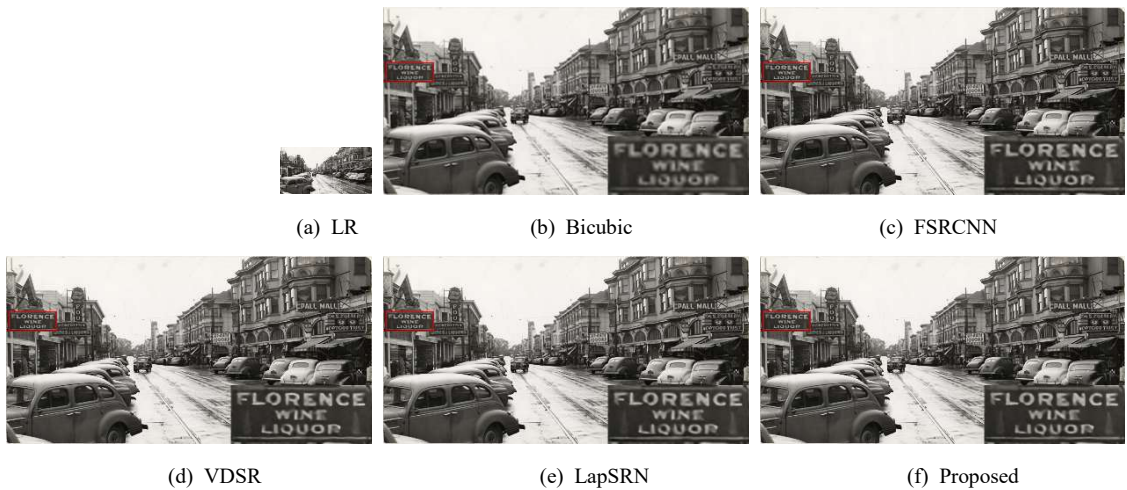

(a) LR          (b) Bicubic          (c) FSRCNN

(d) VDSR          (e) LapSRN          (f) Proposed

**Fig.10** Visual comparison of reconstructed results on real-world historical image by the baseline and proposed methods (scaling factor × 4).

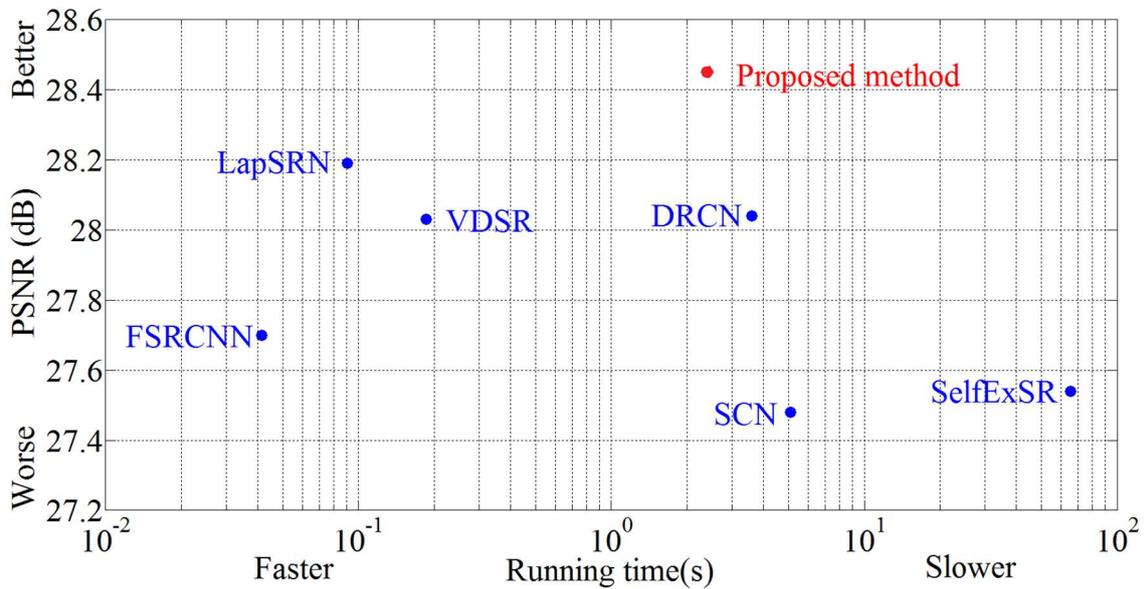

**Fig.11** Average execution time versus PSNR between the proposed and baseline methods. The results are evaluated on all the images in Set14 with the scaling factor of 4. Our method provides the best quality in comparison with other state-of-the-art SR methods and preserves the execution time of SelfExSR, SCN, and DRCN.

**Table 1** Average PSNR and SSIM results of reconstructed HR images by the proposed method and our network with the random initialization method (scaling factor × 3).

| Datasets | Random initialization | | Proposed | |
|---|---|---|---|---|
| | PSNR | SSIM | PSNR | SSIM |
| Set5 | 34.01 | 0.9365 | **34.22** | **0.9431** |
| Set14 | 29.98 | 0.8352 | **30.16** | **0.8374** |

**Table 2** Trade-off between performance and average execution time for the different number of pyramid level in the proposed network. (scaling factor × 3).

| $L$ | Set5 | | Set14 | |
|---|---|---|---|---|
| | PSNR | Second | PSNR | Second |
| 1 | 32.88 | 0.1820 | 29.40 | 0.4616 |
| 2 | 33.36 | 0.4047 | 29.64 | 0.9822 |
| 3 | 33.61 | 0.4354 | 29.76 | 1.0462 |
| 4 | 33.88 | 0.4911 | 29.89 | 1.1596 |
| 5 | 33.97 | 0.5672 | 29.93 | 1.4396 |

**Table 3** Trade-off between performance and average execution time for the different number of inception-residual block on each pyramid level of the proposed network (scaling factor × 3).

| $d$ | Set5 | | Set14 | |
|---|---|---|---|---|
| | PSNR | Second | PSNR | Second |
| 1 | 33.88 | 0.4911 | 29.89 | 1.1596 |
| 2 | 34.13 | 0.7317 | 30.07 | 1.7422 |
| 3 | 34.22 | 0.9625 | 30.16 | 2.2726 |
| 4 | 34.29 | 1.2381 | 30.19 | 2.7915 |

**Table 4** Average PSNR and SSIM for scaling factor ×2 on datasets set5, set14, urban100, manga109 and bsd100 among different methods.

| Methods | Scale | Set5 | Set14 | BSD100 | Urban100 | Manga109 |
|---|---|---|---|---|---|---|
| | | PSNR/SSIM | PSNR/SSIM | PSNR/SSIM | PSNR/SSIM | PSNR/SSIM |
| Bicubic | ×2 | 33.65/0.929 | 30.34/0.868 | 29.56/0.843 | 26.88/0.841 | 30.84/0.935 |
| FSRCNN | ×2 | 36.99/0.955 | 32.73/0.909 | 31.51/0.891 | 29.87/0.901 | 36.62/0.971 |
| SelfExSR | ×2 | 36.49/0.954 | 32.44/0.906 | 31.18/0.886 | 29.54/0.897 | 35.78/0.968 |
| SCN | ×2 | 36.52/0.953 | 32.42/0.904 | 31.24/0.884 | 29.50/0.896 | 35.47/0.966 |
| VDSR | ×2 | 37.53/0.958 | 32.97/0.913 | 31.90/0.896 | 30.77/0.914 | 37.16/0.974 |
| DRCN | ×2 | 37.63/0.959 | 32.98/0.913 | 31.85/0.894 | 30.76/0.913 | 37.57/0.973 |
| LapSRN | ×2 | 37.52/0.959 | 33.08/0.913 | 31.80/0.895 | 30.41/0.910 | 37.27/0.974 |
| Proposed | ×2 | **37.83/0.969** | **33.18/0.924** | **32.01/0.902** | **30.86/0.919** | **37.71/0.986** |

**Table 5** Average PSNR and SSIM for scaling factor ×3 on datasets set5, set14, urban100, manga109 and bsd100 among different methods.

| Methods | Scale | Set5 | Set14 | BSD100 | Urban100 | Manga109 |
| --- | --- | --- | --- | --- | --- | --- |
| | | PSNR/SSIM | PSNR/SSIM | PSNR/SSIM | PSNR/SSIM | PSNR/SSIM |
| Bicubic | ×3 | 30.39/0.868 | 27.55/0.774 | 27.21/0.739 | 24.46/0.735 | 26.96/0.856 |
| FSRCNN | ×3 | 32.71/0.907 | 29.23/0.821 | 28.52/0.790 | 26.42/0.807 | 31.12/0.920 |
| SelfExSR | ×3 | 32.58/0.909 | 29.16/0.820 | 28.29/0.784 | 26.44/0.809 | 30.95/0.918 |
| SCN | ×3 | 33.26/0.917 | 29.55/0.827 | 28.58/0.791 | 26.20/0.808 | 30.25/0.913 |
| VDSR | ×3 | 33.66/0.921 | 29.77/0.831 | 28.82/0.798 | 27.14/0.828 | 32.01/0.933 |
| DRCN | ×3 | 33.82/0.923 | 29.76/0.831 | 28.80/0.796 | 27.15/0.827 | 32.24/0.934 |
| LapSRN | ×3 | - | - | - | - | - |
| Proposed | ×3 | **34.22/0.943** | **30.16/0.847** | **29.21/0.816** | **27.58/0.832** | **32.61/0.949** |

**Table 6** Average PSNR and SSIM for scaling factor ×4 on datasets set5, set14, urban100, manga109 and bsd100 among different methods.

| Methods | Scale | Set5 | Set14 | BSD100 | Urban100 | Manga109 |
| --- | --- | --- | --- | --- | --- | --- |
| | | PSNR/SSIM | PSNR/SSIM | PSNR/SSIM | PSNR/SSIM | PSNR/SSIM |
| Bicubic | ×4 | 28.42/0.810 | 26.10/0.704 | 25.96/0.669 | 23.15/0.659 | 24.92/0.789 |
| FSRCNN | ×4 | 30.71/0.865 | 27.70/0.756 | 26.97/0.714 | 24.61/0.727 | 27.89/0.859 |
| SelfExSR | ×4 | 30.33/0.861 | 27.54/0.756 | 26.84/0.712 | 24.82/0.740 | 27.82/0.865 |
| SCN | ×4 | 30.39/0.862 | 27.48/0.751 | 26.87/0.710 | 24.52/0.725 | 27.39/0.856 |
| VDSR | ×4 | 31.35/0.882 | 28.03/0.770 | 27.29/0.726 | 25.18/0.753 | 28.82/0.886 |
| DRCN | ×4 | 31.53/0.884 | 28.04/0.770 | 27.24/0.724 | 25.14/0.752 | 28.97/0.886 |
| LapSRN | ×4 | 31.54/0.885 | 28.19/0.772 | 27.32/0.728 | 25.21/0.756 | 29.09/0.890 |
| Proposed | ×4 | **31.94/0.902** | **28.49/0.791** | **27.70/0.748** | **25.61/0.776** | **29.49/0.902** |